\documentclass[conference]{IEEEtran}
\IEEEoverridecommandlockouts
\usepackage{latex8}
\usepackage{amsmath,amssymb,amsfonts}
\usepackage{algorithmic}
\usepackage{url}
\usepackage{cite}
\usepackage{graphicx}
\usepackage{textcomp}
\usepackage{xcolor}
\usepackage{multirow}
\usepackage{caption}
\usepackage[shortlabels]{enumitem}
\usepackage[colorlinks=true, linkcolor=blue, urlcolor=blue, citecolor=blue, breaklinks=true]{hyperref}
\begin{document}

\title{OpenRR-5k: A Large-Scale Benchmark for Reflection Removal in the Wild}

\author{
\large{Jie Cai, Kangning Yang, Ling Ouyang, Lan Fu, Jiaming Ding, Jinglin Shen, Zibo Meng}\\ \\
\normalsize{OPPO AI Center, Palo Alto, CA, US}\\
\normalsize{jie.cai@oppo.com}\\
}

\maketitle

\begin{abstract}

Removing reflections is a crucial task in computer vision, with significant applications in photography and image enhancement. Nevertheless, existing methods are constrained by the absence of large-scale, high-quality, and diverse datasets.
In this paper, we present a novel benchmark for Single Image Reflection Removal (SIRR). 
We have developed a large-scale dataset containing 5,300 high-quality, pixel-aligned image pairs, each consisting of a reflection image and its corresponding clean version. Specifically, the dataset is divided into two parts: 5,000 images are used for training, and 300 images are used for validation. Additionally, we have included 100 real-world testing images without ground truth (GT) to further evaluate the practical performance of reflection removal methods.
All image pairs are precisely aligned at the pixel level to guarantee accurate supervision. The dataset encompasses a broad spectrum of real-world scenarios, featuring various lighting conditions, object types, and reflection patterns, and is segmented into training, validation, and test sets to facilitate thorough evaluation.
To validate the usefulness of our dataset, we train a U-Net-based model and evaluate it using five widely-used metrics, including PSNR, SSIM, LPIPS, DISTS, and NIQE. We will release both the dataset and the code on \url{https://github.com/caijie0620/OpenRR-5k} to facilitate future research in this field.

\end{abstract}


\section{Introduction}
\label{sec:introduction}

Single image reflection removal (SIRR) is a vital task in computer vision, with the goal of extracting the clear underlying transmission image from unwanted reflections in a single image.
This task plays a critical role in enhancing image quality across various practical applications, such as photography, autonomous driving~\cite{yang2022online}, augmented reality~\cite{bimber2006modern}, and medical imaging~\cite{shih2023deep}. Current reflection removal techniques range from traditional image decomposition methods to more advanced deep learning-based solutions~\cite{fan2017generic,li2020single,song2023robust,zhong2024language,yang2025ntire,yang2025survey,yang2025openrr1k,cai2025f2t2hit,cai2025vlm,zhao2025reversible}.

Despite notable progress, SIRR remains fundamentally challenging due to the ill-posed nature of the reflection formation process~\cite{wei2019single}. Reflections can differ significantly in intensity, shape, and color, influenced by complex scene geometries and lighting conditions. Early studies typically assumed a simplistic additive model, where an observed image $\mathbf{I}$ is considered a linear combination of a transmission layer $\mathbf{T}$ and reflection layer $\mathbf{R}$, i.e., $\mathbf{I} = \mathbf{T} + \mathbf{R}$~\cite{fan2017generic, levin2007user}. Later approaches refined this model by incorporating blending coefficients~\cite{wan2018crrn, yang2018seeing} or employing alpha-matting mechanisms~\cite{dong2021location} to better approximate real-world conditions.

However, the effectiveness of these methods heavily relies on the availability of high-quality training data, which has become a significant bottleneck. Existing datasets are typically limited in size, diversity, and quality, restricting the development and generalization of data-driven models. To address these issues, we propose a new approach to collecting reflection datasets, focusing explicitly on constructing large-scale, strictly aligned, and diverse image pairs. Our dataset collection protocol places no strict limitations on capture conditions, allowing images with reflections to be sourced flexibly from various real-world scenarios or online platforms, thus ensuring natural diversity.

Crucially, our approach ensures pixel-level alignment between reflection-contaminated images and their clean ground-truth counterparts. Unlike previous methods that remove reflective surfaces physically or use controlled environments~\cite{li2020single, wan2017benchmarking, zhang2018single, lei2020polarized, lei2022categorized, zhu2024revisiting}, we rely on proven reflection removal techniques combined with manual refinement through image editing tools. This approach greatly streamlines the data acquisition process, enhancing its scalability, cost-effectiveness, and suitability for large-scale data collection through crowdsourcing platforms.

Following this protocol, we constructed a new dataset that consists of 5,000 high-quality, strictly pixel-aligned image pairs for training, along with an additional 300 image pairs for validation. These pairs cover diverse real-world scenes and reflection types. We extensively evaluated our dataset using a U-Net-based model and multiple evaluation metrics, including PSNR, SSIM, LPIPS, DISTS, and NIQE. The results demonstrate notable performance improvements and stronger generalization across challenging real-world scenarios.

The main contributions of this paper can be summarized as follows:
\begin{itemize}
\item We propose a novel and scalable data collection protocol to obtain high-quality and pixel-aligned image pairs, significantly improving dataset diversity and realism.
\item We introduce a large-scale dataset comprising 5,300 real-world reflection-contaminated image pairs, strictly aligned at the pixel level, to support robust training and evaluation. Additionally, we provide a real-world testing set of 100 images without ground truth (GT) to further assess the practical performance of models in real-life scenarios.
\item We perform comprehensive benchmark experiments and validate that our dataset effectively enhances the performance and generalization of existing reflection removal datasets.
\end{itemize}

\section{Related Work}
\label{sec:related_work}

Public datasets for single image reflection removal (SIRR) can generally be divided into two main types: fully synthetic and semi-synthetic datasets.

Fully-synthetic datasets are usually generated by merging two reflection-free images through specific blending coefficients to simulate the appearance of reflections. This approach allows the generation of large-scale image pairs efficiently~\cite{fan2017generic, zhang2018single, zhang2023benchmarking}. For instance, Guo et al.\cite{guo2014robust} adopted fixed coefficients, using 0.6 for transmission and 0.4 for reflection layers. Fan et al.\cite{fan2017generic} synthesized reflection images by adaptively combining background and reflection layers, carefully avoiding brightness overflow and applying Gaussian blur to simulate various reflection intensities realistically. Zhang et al.~\cite{zhang2023benchmarking} focused specifically on synthesizing ultra-high-definition reflection images.

Semi-synthetic datasets are constructed using physical setups involving props such as glass panels and black cloths. Researchers typically capture images containing reflections and then physically remove the reflective surface or block reflections with light-absorbing materials, resulting in paired reflection-contaminated and clean transmission images~\cite{li2020single, wan2017benchmarking, zhang2018single, lei2020polarized, lei2022categorized, zhu2024revisiting}. For example, Li et al.\cite{li2020single} captured clean images by manually removing glass surfaces. Lei et al.\cite{lei2020polarized, lei2022categorized} proposed capturing RAW images and extracting transmission layers by subtracting the reflection component. Recently, Zhu et al.~\cite{zhu2024revisiting} presented a more sophisticated pipeline that involves extensively blocking reflections from environmental sources.

Despite their practical utility, existing simulation-based approaches have notable limitations. Fully-synthetic methods heavily rely on simplified assumptions about reflection phenomena, causing significant domain gaps when applied to real-world images~\cite{lei2022categorized}. Meanwhile, semi-synthetic methods often encounter pixel-level misalignment caused by factors such as glass refraction, equipment vibrations, or environmental influences like wind. These issues lead to inconsistencies between paired images. Additionally, blocking reflections using black cloth rarely achieves perfect isolation, resulting in residual reflections and color inconsistencies between the paired images. These limitations significantly restrict the realism, scalability, and diversity of existing datasets. As a result, capturing the natural complexity of real-world reflections—such as their intensity, shape, and color variations under diverse scene geometries and lighting conditions—remains a challenging task. Addressing these issues is crucial for enhancing the performance and robustness of SIRR models in practical applications.

\section{Methodology}
\label{sec:methodology}

\begin{figure*}[!t]
    \centering
    \footnotesize
    \includegraphics[width=0.95\textwidth]{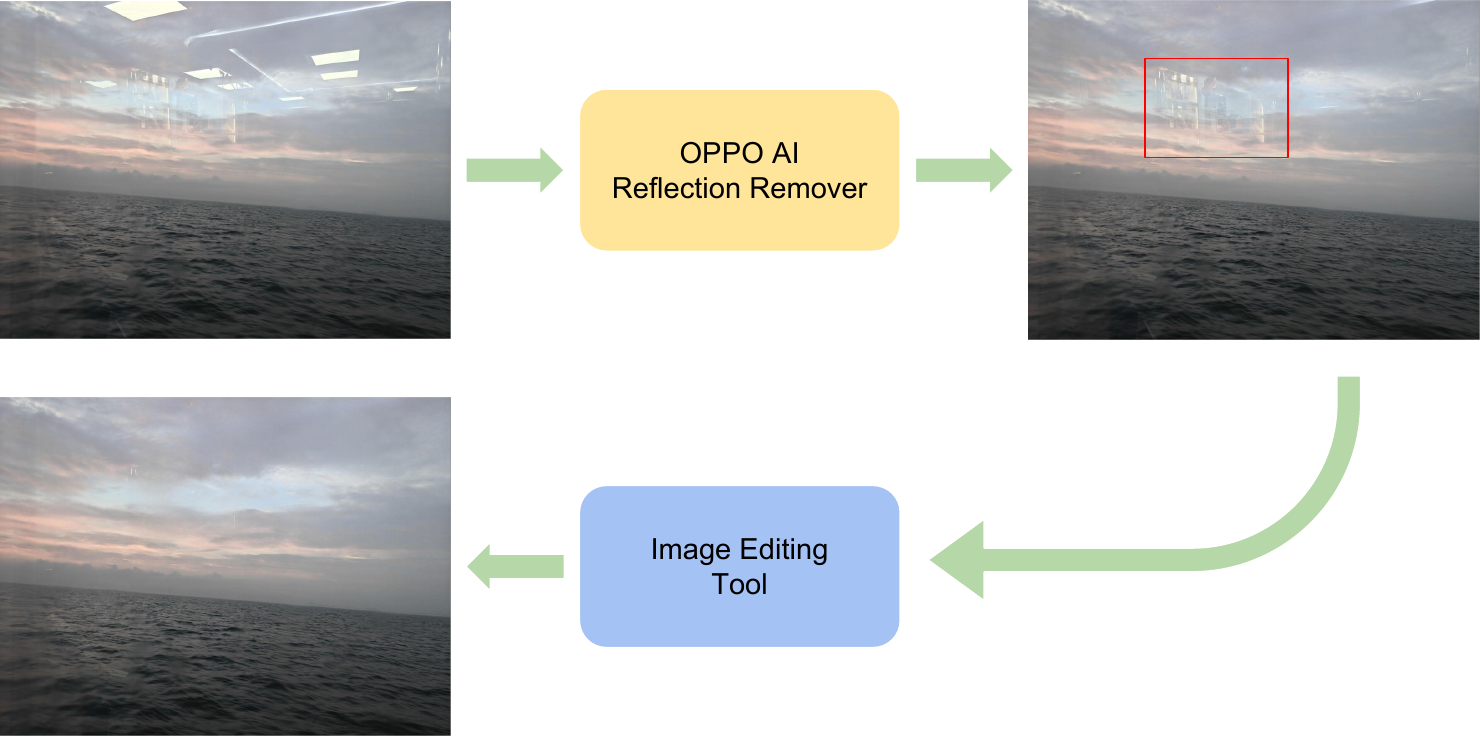}
    \caption{Visualization of paired data generation pipeline for reflection removal.}
    \label{fig:protocol}
\end{figure*}

\subsection{Dataset Collection Protocol}
As shown in Fig.~\ref{fig:protocol}, our proposed data collection protocol consists of two main steps. The first step involves using a proven off-the-shelf tool to initially remove reflections from the images. We adopted the OPPO smartphone's AI-based reflection removal software \footnote{\url{https://www.youtube.com/watch?v=4IUBm18YL68&ab_channel=OPPO}} to obtain the initial reflection removal results.
This commercial software, integrated into OPPO smartphones, is specifically designed to handle reflection artifacts in photographs and is one of the few effective tools currently available on the market for this purpose, with similar tools offered by Samsung AI Reflection Removal.

We observed that the initial reflection removal results removed major reflection components; however, subtle residual reflections remain, as shown in the intermediate image of Fig.~\ref{fig:protocol}. To address this, the second stage of our protocol involves a  refinement process to recover more details.  Specifically, we adopted professional image editing tools (e.g., Photoshop, MeituPic, etc.) for the refinement. This step is crucial for eliminating any remaining artifacts or inconsistencies in the intermediate images. After precise manual adjustments, the final processed images are of high quality and suitable for training and evaluation purposes. This manual intervention allows us to preserve fine details while eliminating any potential artifacts introduced during AI processing.

\begin{figure*}[h]
    \centering
    \footnotesize
    \includegraphics[width=1.0\textwidth]{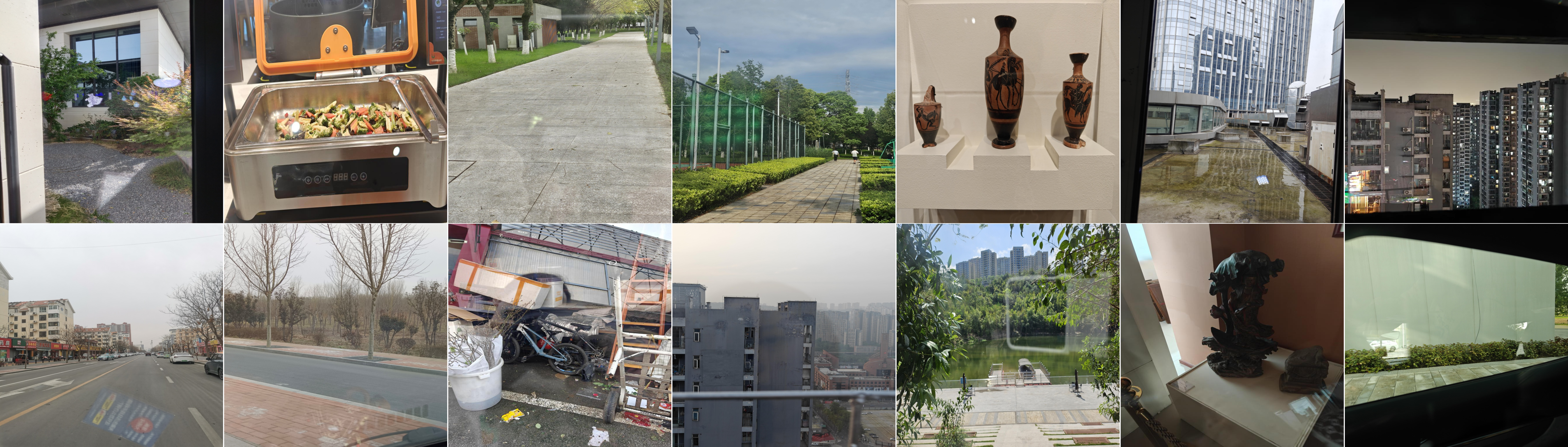}
    \caption{ Overview of our \textit{OpenRR-5k} dataset. }
    \label{fig:dataset_examples} 
\end{figure*}

Compared to existing data collection methods~\cite{wan2017benchmarking, zhang2018single, lei2020polarized, li2020single, lei2022categorized, zhu2024revisiting}, our approach offers several key advantages:

1) \textbf{Diversity}: Our method enables the collection of a significantly broader range of data samples, without being restricted by specific lighting conditions or types of glass surfaces. The collected images cover various real-world reflection scenarios, including diverse lighting conditions (e.g., daylight, sunset, and nighttime illumination) and different glass surfaces, such as car windows, building glass doors, museum display cases, and other types of glass. Notably, this diverse distribution is reflected in our test set, as illustrated in Fig.~\ref{fig:dataset_examples}, which showcases the wide variety of scenarios our method can handle.

2) \textbf{Pixel-level Alignment}: To address the challenges of pixel-level misalignment and inconsistencies in existing datasets, we have employed off-the-shelf tools and techniques to ensure that the input images containing reflections and the corresponding processed transmission images are perfectly aligned at the pixel level. This alignment process is crucial for maintaining consistency and accuracy in the dataset, thereby providing a more reliable foundation for training and evaluating single image reflection removal (SIRR) models. By leveraging these tools, we are able to mitigate the issues associated with factors such as glass refraction, equipment vibrations, and environmental influences, ultimately enhancing the realism and quality of our dataset. 

3) \textbf{True Real-World Data}: Our method eliminates the need for collecting ground-truth data, allowing us to capture authentic reflection scenarios directly from real-world environments. Unlike traditional approaches that rely on artificial setups or synthetic reflections, our technique ensures that the data we collect truly represents genuine real-world situations. This not only enhances the realism and diversity of our dataset but also provides a more accurate basis for training and evaluating single image reflection removal (SIRR) models, ultimately improving their performance and robustness in practical applications.

\subsection{OpenRR-5k Dataset}
Based on our proposed protocol, we constructed the OpenRR-5k dataset, which comprises a total of 5,300 image pairs. Specifically, we allocated 5,000 image pairs for the training set (denoted as OpenRR-5k$_{train}$), 300 image pairs for the validation set (denoted as OpenRR-5k$_{val}$), and 100 image pairs without Ground Truth for the test set (denoted as OpenRR-5k$_{test}$).

\begin{table*}[ht]
\centering
\normalsize
\caption{Comparison of Existing Datasets with Our OpenRR-5k Dataset}
\label{tab:1}
\begin{tabular}{c|c|c|c|c}
Dataset & Year & Usage      & \begin{tabular}[c]{@{}c@{}}Pair \\ Number\end{tabular} & \begin{tabular}[c]{@{}c@{}}Average \\ Resolution\end{tabular} \\ \hline
SIR$^2$~\cite{wan2017benchmarking}   & 2017 & Test       & 454         & 540 x 400    \\
Real~\cite{zhang2018single}        & 2018 & Train/Test & 89/20       & 1152 x 930     \\
Nature~\cite{li2020single}      & 2020 & Train/Test & 200/20      & 598 x 398         \\
RRW~\cite{zhu2024revisiting}      & 2023 & Train & 14952      & 2580 × 1460           \\
OpenRR-1k~\cite{yang2025openrr1k}        & 2025 & Train/Val/Test & 800/100/100    &  922 x 917                \\
\hline
\textbf{OpenRR-5k}    & 2025 & Train/Val/Test & 5,000/300/100    &  874 x 931         \\ 
\end{tabular}
\end{table*}

Table~\ref{tab:1} presents a comprehensive comparison between our OpenRR-5k dataset and other publicly available reflection removal datasets. Compared to SIR$^2$\cite{wan2017benchmarking}, Real\cite{zhang2018single}, and Nature~\cite{li2020single}, our OpenRR-5k dataset includes more data samples and higher image resolution. Although RRW~\cite{zhu2024revisiting} contains more data pairs and higher resolution images, we argue that our dataset offers higher-quality samples and greater convenience. Specifically, our dataset does not require specialized data collection equipment or consideration of various environmental factors, making it more accessible and practical for a wider range of users. In fact, when attempting to use the RRW pipeline, we found that it was difficult to operate in practical deployments due to the complexities involved in setting up and maintaining the required equipment and conditions.
In addition to OpenRR-1k~\cite{yang2025ntire}, we extend the dataset to a larger scale, termed OpenRR-5k. The OpenRR-1k dataset consists of 800 training, 100 validation, and 100 test image pairs. We include all 1k image pairs from OpenRR-1k into the training set of OpenRR-5k. In addition, we collect 300 new validation image pairs and 100 test reflection images without ground truth.

\begin{figure*}[h]
    \centering
    \footnotesize
    \includegraphics[width=1.0\textwidth]{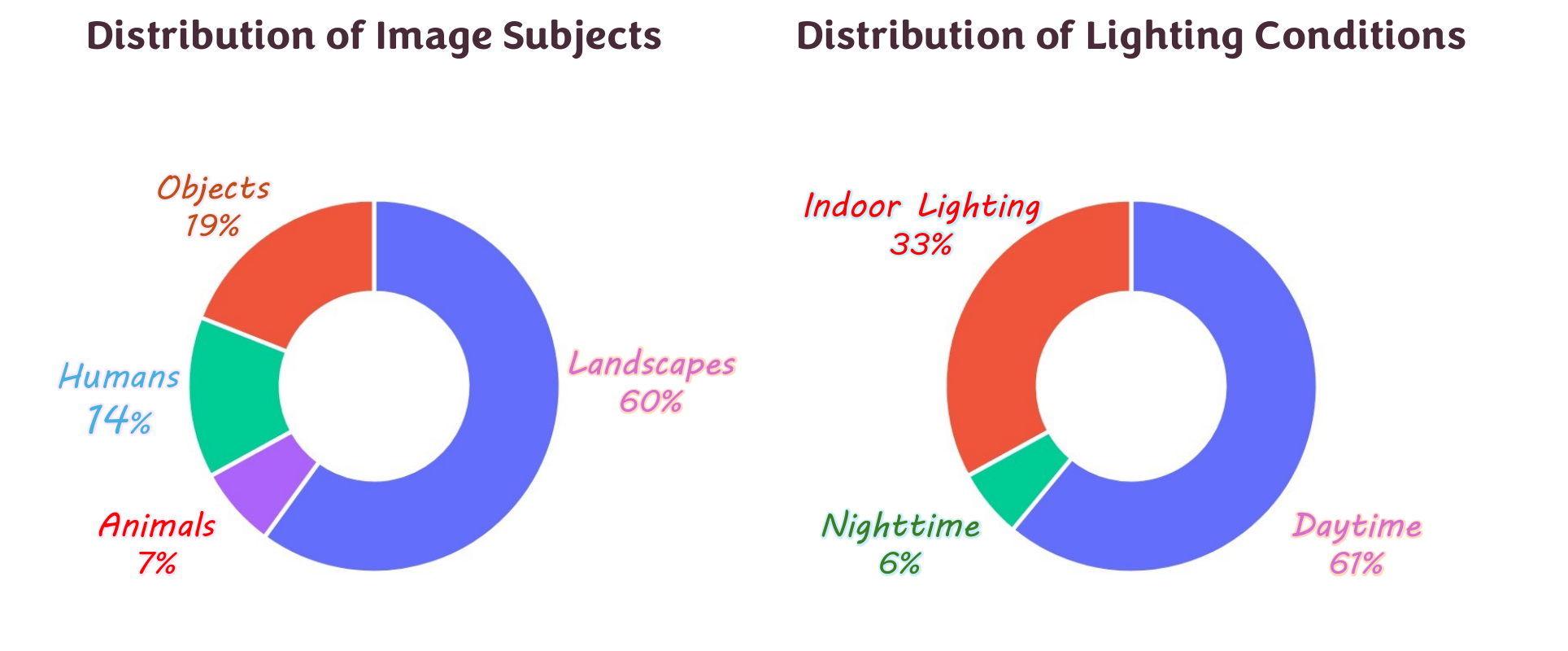}
    \caption{The category distribution of our \textit{OpenRR-5k$_{\text{test}}$} dataset}
    \label{fig:statistics}	 
\end{figure*}

Additionally, Fig.~\ref{fig:statistics} offers a detailed overview of the categorical composition of our OpenRR-5k dataset, specifically focusing on the test set. The distribution is analyzed from two key perspectives: scene content and lighting conditions. For scene content (illustrated in the left pie chart), the test set is categorized into five main groups: humans, animals, inanimate objects, and urban/natural landscapes. In terms of lighting conditions (depicted in the right pie chart), the test set is divided across three distinct scenarios: daytime, nighttime, and indoor lighting. This diverse distribution ensures that our test set covers a wide range of real-world reflection scenarios, making it a comprehensive benchmark for evaluating the robustness and generalization of single image reflection removal models.

\section{Experiments}
\label{sec:experiments}

\begin{table*}[htbp]
\centering
\normalsize
\caption{Quantitative Comparisons of Real-World Reflection Removal Datasets}
\label{tab:2}
\begin{tabular}{l|c|c|c|c}
\textbf{Method} & \textit{Nature} (20) & \textit{Real} (20) & \textit{SIR$^2$} (454) & \textit{OpenRR-5k$_{\text{val}}$} (300) \\ \hline
NAFNet    &  25.62   &  21.16   &    24.52      &    26.59     \\
\end{tabular}
\end{table*}

\begin{table*}[htbp]
\centering
\normalsize
\caption{Comprehensive Quantitative Comparisons on \textit{OpenRR-5k$_{\text{val}}$}}
\label{tab:3}
\begin{tabular}{l|c|c|c|c|c}
\textbf{Metrics} & \textit{PSNR} & \textit{SSIM} & \textit{LPIPS} & \textit{DISTS} & \textit{NIQE} \\ \hline
NAFNet &  26.59   &  0.9418   &    0.0911      &    0.0538   &   3.4066    \\
\end{tabular}
\end{table*}

\subsection{Experiment setting}
To conduct a comprehensive benchmark evaluation on the proposed OpenRR-5k dataset, we developed a new baseline model based on the NAFNet architecture, adapting the widely-used restoration framework introduced in~\cite{chen2022simple}. To enhance the model's representation learning capabilities, we expanded the network's bottleneck capacity by increasing the number of encoder blocks, middle blocks, and decoder blocks from 1 to 2, resulting in 2 blocks for each of the components. This increase in depth enables the model to process global image features more effectively, thereby improving its ability to capture and handle complex reflection patterns. 

We trained the NAFNet model directly on the OpenRR-5k training dataset to assess whether the proposed dataset could enhance the model's generalization ability. Subsequently, as shown in Table~\ref{tab:2}, we evaluated the trained model on the validation sets of \textit{Nature}, \textit{Real}, \textit{SIR$^2$}, and \textit{OpenRR-5k\_val}, using the Peak Signal-to-Noise Ratio (PSNR) as the key evaluation metric. The PSNR values were calculated in the RGB color space, with higher scores indicating better performance. This simple training and validation process enabled us to rapidly assess the effectiveness of the OpenRR-5k dataset and the NAFNet model in dealing with a variety of image data. In addition to the PSNR values, we also reported results on four other evaluation metrics, namely SSIM, LPIPS, DISTS, and NIQE, as detailed in Table~\ref{tab:3}.

\subsection{Implementation details}
Our framework is implemented with the PyTorch platform. During the training phase, the network is trained using the Adam optimizer with an initial learning rate of 0.0001, which is adjusted based on a Cosine Annealing Restart scheme. The scheduler is configured with three periods of 100,000 iterations each and corresponding restart weights of 1, 0.5, and 0.25. The total number of iterations is set to 300,000. The training is conducted with eight Nvidia A100 GPUs for approximately 24 hours. The batch size per GPU is set to 1, and 512 × 512 patches are randomly cropped from the images at each training iteration. Data augmentation includes random horizontal flipping and random rotation.

\section{Conclusion}
\label{sec:conclusion}

In this paper, we propose a novel reflection removal pipeline that addresses key challenges in single image reflection removal (SIRR). Traditional methods are often limited by the difficulty of collecting diverse, high-quality real-world data. Our pipeline provides a more accessible and cost-effective way to gather such data, enabling the construction of the OpenRR-5k dataset, which includes 5,000 training image pairs, 300 validation image pairs, and 100 test images without ground truth. This dataset covers a wide range of real-world scenarios, including different lighting conditions and types of glass surfaces.

To demonstrate the value of OpenRR-5k, we adapt a NAFNet-based baseline model to better fit the dataset’s characteristics. Benchmark results show notable performance improvements over existing methods, even though the model is trained exclusively on our OpenRR-5k dataset without using any additional training data. This highlights the effectiveness of our dataset in enhancing current SIRR models and its potential to support more robust and practical reflection removal solutions.

{
\bibliographystyle{latex8}
\bibliography{reference}

\begin{thebibliography}{10}\setlength{\itemsep}{1ex}\normalsize

\bibitem{yang2022online}
J.~Yang, H.~Ge, J.~Yang, Y.~Tong, and S.~Su,
\newblock ``Online multi-object tracking using multi-function integration and tracking simulation training,''
\newblock {\em Applied Intelligence}, 2022.

\bibitem{bimber2006modern}
O.~Bimber and R.~Raskar,
\newblock ``Modern approaches to augmented reality,''
\newblock in {\em ACM SIGGRAPH 2006 Courses}, 2006.

\bibitem{shih2023deep}
Chi-Sheng Shih, Yu-Cheng Liao, and Ching-Ting Tan,
\newblock ``Deep learning based end-to-end specular reflection removal for medical endoscopic images,''
\newblock in {\em Proceedings of the 2023 International Conference on Research in Adaptive and Convergent Systems}, 2023.

\bibitem{fan2017generic}
Q.~Fan, J.~Yang, G.~Hua, B.~Chen, and D.~Wipf,
\newblock ``A generic deep architecture for single image reflection removal and image smoothing,''
\newblock in {\em ICCV}, 2017.

\bibitem{li2020single}
C.~Li, Y.~Yang, K.~He, S.~Lin, and J.~E Hopcroft,
\newblock ``Single image reflection removal through cascaded refinement,''
\newblock in {\em CVPR}, 2020.

\bibitem{song2023robust}
Z.~Song, Z.~Zhang, K.~Zhang, W.~Luo, Z.~Fan, W.~Ren, and J.~Lu,
\newblock ``Robust single image reflection removal against adversarial attacks,''
\newblock in {\em CVPR}, 2023.

\bibitem{zhong2024language}
H.~Zhong, Y.~Hong, S.~Weng, J.~Liang, and B.~Shi,
\newblock ``Language-guided image reflection separation,''
\newblock in {\em CVPR}, 2024, pp. 24913--24922.

\bibitem{yang2025ntire}
K.~Yang, J.~Cai, L.~Ouyang, F.~Vasluianu, R.~Timofte, et~al.,
\newblock ``Ntire 2025 challenge on single image reflection removal in the wild: Datasets, methods and results,''
\newblock in {\em CVPR Workshops}, 2025.

\bibitem{yang2025survey}
K.~Yang, H.~Sun, J.~Cai, L.~Fu, J.~Ding, J.~Li, and Z.~Meng,
\newblock ``Survey on single-image reflection removal using deep learning techniques,''
\newblock in {\em MIPR}, 2025.

\bibitem{yang2025openrr1k}
K.~Yang, L.~Ouyang, H.~Sun, J.~Cai, L.~Fu, J.~Ding, C.~M. Ho, and Z.~Meng,
\newblock ``Openrr-1k: A scalable dataset for real-world reflection removal,''
\newblock in {\em ICIP}, 2025.

\bibitem{cai2025f2t2hit}
J.~Cai, K.~Yang, L.~Ouyang, L.~Fu, J.~Ding, H.~Sun, C.~M. Ho, and Z.~Meng,
\newblock ``F2t2-hit: A u-shaped fft transformer and hierarchical transformer for reflection removal,''
\newblock in {\em ICIP}, 2025.

\bibitem{cai2025vlm}
J.~Cai, K.~Yang, J.~Ding, L.~Fu, L.~Ouyang, J.~Li, J.~Shen, and Z.~Meng,
\newblock ``Degradation-aware image enhancement via vision-language classification,''
\newblock in {\em MIPR}, 2025.

\bibitem{zhao2025reversible}
H.~Zhao, M.~Li, Q.~Hu, and X.~Guo,
\newblock ``Reversible decoupling network for single image reflection removal,''
\newblock in {\em CVPR}, 2025.

\bibitem{wei2019single}
K.~Wei, J.~Yang, Y.~Fu, D.~Wipf, and H.~Huang,
\newblock ``Single image reflection removal exploiting misaligned training data and network enhancements,''
\newblock in {\em CVPR}, 2019.

\bibitem{levin2007user}
Anat Levin and Yair Weiss,
\newblock ``User assisted separation of reflections from a single image using a sparsity prior,''
\newblock {\em IEEE Transactions on Pattern Analysis and Machine Intelligence}, 2007.

\bibitem{wan2018crrn}
Renjie Wan, Boxin Shi, Ling-Yu Duan, Ah-Hwee Tan, and Alex~C Kot,
\newblock ``Crrn: Multi-scale guided concurrent reflection removal network,''
\newblock in {\em CVPR}, 2018.

\bibitem{yang2018seeing}
Jie Yang, Dong Gong, Lingqiao Liu, and Qinfeng Shi,
\newblock ``Seeing deeply and bidirectionally: A deep learning approach for single image reflection removal,''
\newblock in {\em ECCV}, 2018.

\bibitem{dong2021location}
Zheng Dong, Ke~Xu, Yin Yang, Hujun Bao, Weiwei Xu, and Rynson~WH Lau,
\newblock ``Location-aware single image reflection removal,''
\newblock in {\em ICCV}, 2021.

\bibitem{wan2017benchmarking}
Renjie Wan, Boxin Shi, Ling-Yu Duan, Ah-Hwee Tan, and Alex~C Kot,
\newblock ``Benchmarking single-image reflection removal algorithms,''
\newblock in {\em ICCV}, 2017, pp. 3922--3930.

\bibitem{zhang2018single}
Xuaner Zhang, Ren Ng, and Qifeng Chen,
\newblock ``Single image reflection separation with perceptual losses,''
\newblock in {\em Proceedings of the IEEE conference on computer vision and pattern recognition}, 2018.

\bibitem{lei2020polarized}
Chenyang Lei, Xuhua Huang, Mengdi Zhang, Qiong Yan, Wenxiu Sun, and Qifeng Chen,
\newblock ``Polarized reflection removal with perfect alignment in the wild,''
\newblock in {\em CVPR}, 2020, pp. 1750--1758.

\bibitem{lei2022categorized}
Chenyang Lei, Xuhua Huang, Chenyang Qi, Yankun Zhao, Wenxiu Sun, Qiong Yan, and Qifeng Chen,
\newblock ``A categorized reflection removal dataset with diverse real-world scenes,''
\newblock in {\em CVPR}, 2022, pp. 3040--3048.

\bibitem{zhu2024revisiting}
Y.~Zhu, X.~Fu, Peng-Tao Jiang, H.~Zhang, Q.~Sun, J.~Chen, Zheng-Jun Zha, and B.~Li,
\newblock ``Revisiting single image reflection removal in the wild,''
\newblock in {\em CVPR}, 2024.

\bibitem{zhang2023benchmarking}
Zhenyuan Zhang, Zhenbo Song, Kaihao Zhang, Zhaoxin Fan, and Jianfeng Lu,
\newblock ``Benchmarking ultra-high-definition image reflection removal,''
\newblock {\em arXiv preprint arXiv:2308.00265}, 2023.

\bibitem{guo2014robust}
Xiaojie Guo, Xiaochun Cao, and Yi~Ma,
\newblock ``Robust separation of reflection from multiple images,''
\newblock in {\em CVPR}, 2014, pp. 2187--2194.

\bibitem{chen2022simple}
L.~Chen, X.~Chu, X.~Zhang, and J.~Sun,
\newblock ``Simple baselines for image restoration,''
\newblock in {\em ECCV}, 2022.

\end{thebibliography}
}

\end{document}